\pdfoutput=1

\documentclass[11pt]{article}

\usepackage{naacl2021}

\usepackage{times}
\usepackage{latexsym}
\usepackage{epsfig}
\usepackage{graphicx}
\usepackage{amsmath}
\usepackage{amssymb}
\usepackage[shortlabels]{enumitem}
\usepackage{booktabs}
\usepackage{subcaption}
\usepackage{tabularx}
\usepackage{csquotes}

\MakeOuterQuote{"}
\usepackage{sidecap}
\usepackage[none]{hyphenat}

\newcommand{\etal}{\textit{et al.}}

\newcommand{\eg}{\textit{e.g.~}}
\newcommand{\etc}{\textit{etc.~}}
\newcommand{\cf}{\textit{cf.}}

\usepackage{verbatim}

\newcolumntype{b}{>{\small}X}
\newcolumntype{s}{>{\small\hsize=.24\hsize}X}

\usepackage[T1]{fontenc}

\usepackage[utf8]{inputenc}

\usepackage{microtype}

\title{SynthRef: Generation of Synthetic Referring Expressions \\for Object Segmentation}

\author{Ioannis Kazakos$^{1,2}$, Carles Ventura$^3$, Míriam Bellver$^{1,4}$, Carina Silberer$^5$, Xavier Giró-i-Nieto$^{1,4}$  \\
   \small$^1$\emph{Universitat Politècnica de Catalunya}\quad
   \small$^2$\emph{National Technical University of Athens}\quad
    \small$^3$\emph{Universitat Oberta de Catalunya}\\
   \small$^4$\emph{Barcelona Supercomputing Center}\quad  
   \small$^5$\emph{University of Stuttgart}\\
}

\begin{document}
\maketitle
\begin{abstract}
Recent advances in deep learning have brought significant progress 
in visual grounding tasks such as language-guided video object segmentation. However, collecting large datasets for these tasks is expensive in terms of annotation time, which represents a bottleneck. To this end, 
we propose a novel method, namely SynthRef, for generating synthetic referring expressions for target objects in an image (or video frame), and we also present and disseminate the first large-scale dataset with synthetic referring expressions for video object segmentation. Our experiments demonstrate that by training with our synthetic referring expressions one can improve the ability of a model to generalize across different datasets, without any additional annotation cost. Moreover, our formulation allows its application to any object detection or segmentation dataset. \noindent Project site: \url{https://imatge-upc.github.io/synthref/}
\end{abstract}

\section{Introduction}
\label{sec:introduction}


Visual grounding tasks provide challenging benchmarks for 
artificial intelligence systems, as they must combine vision and language effectively. 
Among them, we focus on 
referring video object segmentation, in which a language query defines which instance to segment from a video sequence. 
In particular, we define \textit{referring expressions (REs)} as linguistic phrases that allow the unique identification of an individual object~(the \textsl{referent}) in a discourse or scene. 
 (\cf,~\citealt{reiter1992fast,qiao2020referring}).
One of the biggest challenges for this task is the lack of relatively large annotated datasets since a tremendous amount of time and human effort is required for annotation.

Using referring expressions to identify objects in the real world lies at the core of human communication. 
Their use for segmenting objects in images has been previously addressed \citep{hu2016segmentation, liu2017recurrent, yu2018mattnet, ye2019cross, chen2019see} and has benefited from large scale datasets, such as  RefCOCO~\citep{kazemzadeh2014referitgame}. 
However, fewer works have explored the segmentation of objects using REs in the video domain, 
although this provides the more natural setup compared to the image domain. 
Humans use referring expressions to identify objects for others in a moving world, better represented by videos than by still images.
\citet{khoreva2018video} were the first to transfer the referring expression segmentation task from images to videos by collecting referring expressions for the DAVIS-2017~\citep{Pont-Tuset_arXiv_2017} dataset. 
%
Later \citet{gavrilyuk2018actor} provided natural language descriptions as guidance for actor segmentation in A2D~\citep{XuHsXiCVPR2015} and J-HMDB~\citep{Jhuang:ICCV:2013}, two datasets used for action and human pose recognition and segmentation. Finally, the first large-scale benchmark for referring video object segmentation, Refer-YouTube-VOS, was built by \citet{seourvos} on top of YouTube-VOS~\citep{xu2018youtube}, a benchmark for video object segmentation.
%

%

\begin{figure*}[t!]
\begin{center}
\includegraphics[width=\linewidth]{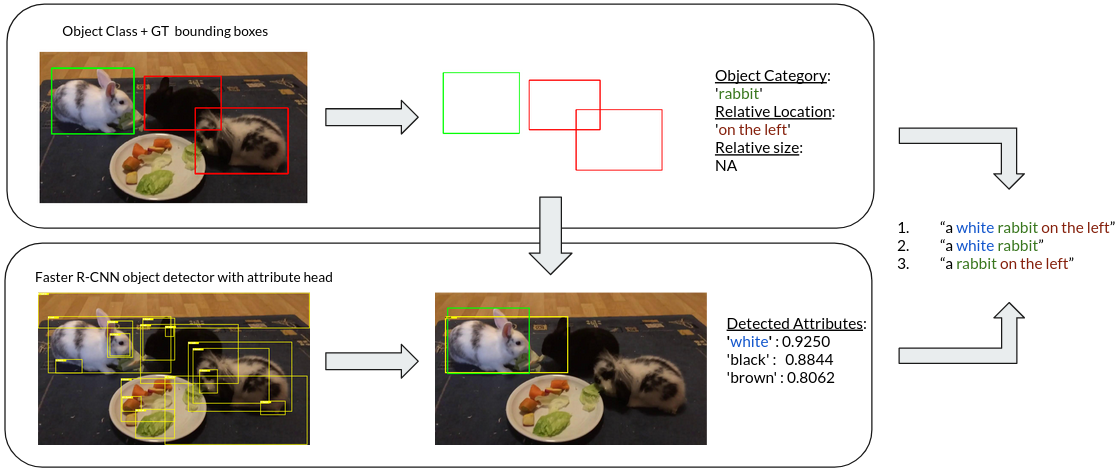}
\end{center}
\caption{Overview of our method for generating synthetic referring expressions. Top: Ground truth labels (object class + bounding boxes) are used to compute a target object's relative location and size. Bottom: A Faster R-CNN object detector with attribute head predicts visual attributes for the detected objects, which are filtered by ground truth annotations. The combined cues create a set of referring expressions that uniquely describe the target object.}
\label{fig:method}
\end{figure*}

As an alternative to collecting REs from annotators, we propose generating synthetic referring expressions for an image, using only the ground truth annotations of objects and their predicted visual attributes from an off-the-shelf deep learning model. 
We apply this method to build a large-scale dataset with synthetic referring expressions for video object segmentation, based on an existing benchmark dataset for video instance segmentation. We use our synthetic dataset for pretraining a deep neural network for the task of referring video object segmentation and evaluate our method on two benchmark datasets used in language-guided video object segmentation.

\section{Method and Dataset}
\label{sec:method}




Our synthetic referring expressions are based on the ground-truth annotations of YouTube-VIS~\citep{yang2019video} dataset, described in the supplementary material. Specifically, we use the classes and bounding boxes of the target and other objects in a video frame, to determine a set of cues 
from which we heuristically generate 
a referring expression that is close to a natural language expression. We call our approach 
\textit{SynthRef}, illustrated in Figure~\ref{fig:method}. 
We use the following four cues for generating referring expressions for a target object:

\begin{table*}[t!]
\resizebox{\linewidth}{!}{
\centering
\setlength{\arrayrulewidth}{1mm}
\setlength{\tabcolsep}{7pt}
\small
\begin{tabular}{@{}lccccc@{}}
\toprule
\textbf{Dataset} & \textbf{Videos / Objects / Categories} & \textbf{Expressions}  & \textbf{Expressions} & \textbf{Expressions/Object} & \textbf{Words/Expression}\\ 
\midrule

    A2D Sentences  &  3,782 / 4,825 / 8 & \textit{Human}  & 6,656  & 1.4 & 7.3  \\
    
    DAVIS-2017  &  150 / 386 / 78 & \textit{Human}  & 1,544 & 4.0 & 5.5\\
   
    Refer-YouTube-VOS  &  3,975 / 7,451 / 94 & \textit{Human}  & 27,899 & 3.7 & 7.5 \\

    SynthRef-YouTube-VIS &   2,238 / 3,774 / 40 & \textit{Synthetic} & 15,798 & 4.2 & 4.4\\

\bottomrule
\end{tabular}
}
\caption{Statistics of our dataset and comparison to existing ones. The last two columns represent the average number of unique referring expressions per object and the average number of words per referring expression respectively.}
\label{table:statistics}
\end{table*}

\noindent
\textbf{1. Object class}
\quad 
In trivial cases where a single object of a known class is present, using the object class is enough to generate a referring expression. However, most cases involve multiple objects of the same class, thus other cues are necessary in order to disambiguate between instances.

\noindent
\textbf{2. Relative size}\quad 
If the total area of the bounding box of the referent is twice as big/small than the area(s) of the respective bounding boxes of the other object(s) of the same class, a characterization of \textit{"bigger/smaller"} or \textit{"the biggest/smallest"} is added to the synthetic referring expression, \eg,~\textit{"the smallest dog"}.

\noindent
\textbf{3. Relative location}\quad In scenarios where two or three objects of the same class are present in a video frame, relative location 
between these objects may suffice 
to disambiguate between them. If the bounding boxes of the objects are fully separable, or partially above a certain threshold, then we assume that relative location of the referent with respect to the other object(s) of the same class can be used in order to generate a non-ambiguous referring phrase. In this case, the steps for determining relative location are the following:

\begin{enumerate}[noitemsep]
    \item The axis which is the most separative for the bounding boxes of the two objects is determined.
    \item According to the axis found and the position of the bounding boxes, a relative location description is given out of 4 options: \{\textit{"on the right", "on the left", "in the back", "in the front"}\}.
    \item If there are two other objects of the same class, steps 1 \& 2 are computed between the referent and each of the two other objects, and the results are combined, \eg,~\textit{"in the middle", "in the back right"},~\etc.
\end{enumerate}

\noindent
\textbf{4. Attributes}\quad 
We pretrain Faster R-CNN~\citep{ren2015faster} on Visual Genome~\citep{krishna2017visual} for object and attribute detection~\citep{tang2020unbiased}. 
This model analyzed the video frames of YouTube-VIS~\citep{yang2019video} dataset to obtain, for each frame of a video, a set of detected objects (with their bounding box coordinates) and their predicted attributes. The detected bounding box with the highest overlap, in terms of Intersection-over-Union (IoU), with the ground truth bounding box of the target object is considered as the prediction corresponding to the referent, as long as its IoU is over 50\%. 
Figure~\ref{fig:method} shows the full pipeline: 
\citeauthor{tang2020unbiased}'s~(\citeyear{tang2020unbiased}) model can detect a total of 201 attributes, which 
 we group to color-like and not color-like attributes, where the latter can be both adjectives (\eg,~\textit{"large", "spotted"}) or verbs (\eg,~\textit{"walking", "surfing"}). 
 The ones with the highest prediction score, if above a certain threshold, are selected for the two subsets, while combinations of two colors are also possible if their scores are very close (\eg,~\textit{"a yellow and green parrot"}). 
 We add an attribute to the referring expression 
only if no other objects belonging to the same class share the same attribute, so that the expression is able to disambiguate between instances. 

\begin{figure*}[t!]
\centering
\begin{subfigure}[t]{0.49\textwidth}
\centering
\includegraphics[width=\textwidth]{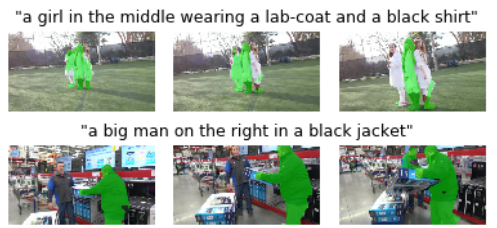}
\caption{Pretrained on RefCOCO}
\label{subfig:refcoco}
\end{subfigure}
\hfill
\begin{subfigure}[t]{0.49\textwidth}
\centering
\includegraphics[width=\textwidth]{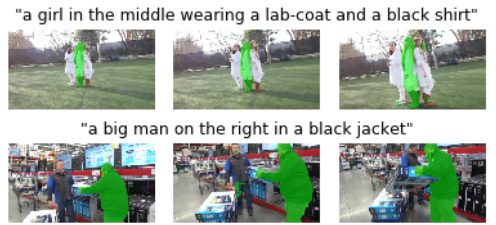}
\caption{Pretrained on RefCOCO+SynthRef-YouTube-VIS}
\label{subfig:ours}
\end{subfigure}
\caption{Qualitative results on DAVIS-2017. Subfigure \ref{subfig:refcoco} (left) shows results when the model is pretrained only on RefCOCO, while Subfigure \ref{subfig:ours} (right) when it is also trained on our synthetic dataset.}
\label{fig:results}
\end{figure*}

Finally, we combine the aforementioned components in a natural order and add a proper article to the sentence, ending up with a synthetic referring expression. 
There might be cases where the generated synthetic language expression may be ambiguous, especially in cases of many similar objects of the same class, although in most cases the generated expression uniquely identifies the referent. 
SynthRef treats each video frame separately, so we do not force any temporal coherence of the REs.
Actually, since 
an object may change its location or appearance throughout the video, we generate one or more synthetic referring expressions for each frame of the video. 
%
In this way, a model can be trained with different referring expressions for the same video or frame increasing its ability to generalize.

Basic statistics of our \textit{SynthRef-YouTube-VIS} dataset and a comparison with other relevant ones are presented in Table \ref{table:statistics}. 
The comparison shows that our dataset, despite not being the largest one in terms of number of annotated objects and their categories, it still has the highest average number of unique referring expressions per annotated object (4.2) without involving any human annotation cost. The average number of words in our referring expressions is 4.4, which is smaller than 
those of the other datasets as our goal is to generate simple and efficient synthetic referring expressions.
We point out three limitations of our dataset/method: (a) the predicted attributes may be wrong or not disambiguating, (b) the relative location is not applied for more than three objects of the same class, and (c) when none of our rules can be applied, SynthRef uses just the object class (\eg ~\textit{"a dog"}), even if there are more instances of that class.

\section{Experiments}

\label{sec:experiments}


We show the benefits of our synthetic dataset SynthRef-YouTube-VIS by using it in extending the training dataset of RefVOS~\citep{bellver2020refvos}, a state of the art model for referring video object segmentation.
The first experiments focus on DAVIS-2017~\citep{Pont-Tuset_arXiv_2017}, and the latter on Refer-YouTube-VOS~\citep{seourvos}.
\begin{table}[ht]
\centering
\begin{tabularx}{\linewidth}{bbbs}
\toprule
RefCOCO  &  SynthRef & DAVIS-2017 & J\&F$\uparrow$     \\ 
\midrule
$\checkmark$  &    &   val & 40.8 \\
$\checkmark$  &   $\checkmark$  &  val & \textbf{44.8} \\
\midrule
$\checkmark$  &    &   train+val & 33.6 \\
                &  $\checkmark$  &   train+val & 27.0 \\
$\checkmark$  &   $\checkmark$  &  train+val & \textbf{38.6} \\
\bottomrule
\end{tabularx}
\caption{Segmentation accuracy obtained with RefVOS model on two partitionns of DAVIS-2017: validation (\textit{val}) or training+validation (\textit{train+val}). Adding our SynthRef-YouTube-VIS data significantly increases the performance at a zero-cost in annotation. The J\&F metric is defined in the supplementary material.}
\label{table:davis-nofinetune}
\end{table}

\begin{table}[h]
\centering
\begin{tabularx}{\linewidth}{bss}
\toprule
Model & SynthRef & J\&F$\uparrow$     \\ 
\midrule
    \citep{khoreva2018video}   &  & 39.3         \\
    \citep{seourvos} &   & 44.1         \\
    \citep{bellver2020refvos}  &   &    45.1 \\
    \citep{bellver2020refvos}  &  $\checkmark$            &    \textbf{45.3}     \\
\bottomrule
\end{tabularx}
\caption{Comparison with the state of the art in DAVIS-2017 validation, with models pretrained on RefCOCO and fine-tuned with DAVIS-2017 training data. Adding our generated SynthRef-YouTube-VIS dataset to the RefCOCO pretraining achieves state of the art results. However the relative gain is smaller than in the scenario without fine-tuning, reported in Table \ref{table:davis-nofinetune}.}
\label{table:davis-finetune}
\end{table}

\noindent
\textbf{DAVIS-2017}
\quad
We report the gains of adding the synthetic dataset when evaluating on the standard validation partition of DAVIS-2017 (30 videos), but also on the combined training and validation partitions (90 videos), to obtain more statistically significant results.
The results in Table~\ref{table:davis-nofinetune} show a significant improvement in segmentation accuracy when adding our synthetic REs to RefCOCO. 
Figure~\ref{fig:results} illustrates qualitative results for this scenario, where the improvement of the segmentation masks is cleary visible. 
The gain is minor when RefVOS is fine-tuned with training data from DAVIS-2017 and evaluated on the validation partition, as shown in Table~\ref{table:davis-finetune}.
This setup is the commonly adopted by the related work, allowing a comparison of our results with them.

\begin{table}[h]
\resizebox{\linewidth}{!}{
\begin{tabular}{lccc}
\toprule
RE Source & Prec@0.5$\uparrow$ & Prec@0.9$\uparrow$ & Mean IoU$\uparrow$       \\ 
\midrule
Synthetic   & 32.3  & 1.8  & 35.0           \\
Human  &   \textbf{38.6}  &  \textbf{6.9}  &  \textbf{39.5}          \\  
\bottomrule
\end{tabular}
}
\caption{Comparison of the performance on a subset of Refer-YouTube-VOS when training with synthetic and human referring expressions.}
\label{table:synth-real}
\end{table}



\noindent
\textbf{Refer-YouTube-VOS}
\quad We further evaluate our method using the subset of Refer-YouTube-VOS that corresponds to our synthetic dataset, SynthRef-YouTube-VIS. We train two instances of RefVOS, one using the 
human-produced REs of Refer-YouTube-VOS and one using the synthetic REs of SynthRef-YouTube-VIS. The evaluation is done on the test split of SynthRef-YouTube-VIS but using the human REs of Refer-YouTube-VOS in both models for a fair comparison. Since both human and synthetic expressions (ours) are available for the same videos, we can measure the domain gap between real and synthetic data for training. Our results, reported in Table~\ref{table:synth-real}, indicate that, even though the model trained on human referring expressions outperforms the model trained on synthetic ones, the drop in accuracy is not that big to prevent the use of our synthetic data for training. On the contrary, the obtained numbers show that our synthetic expressions can be used interchangeably with the human ones when the latter are hard to acquire.


\section{Conclusion}
\label{sec:conclusion}

In this work, we propose SynthRef, a novel method for generating synthetic referring expressions, which is used to create the first large-scale dataset of synthetic referring expressions for video object segmentation, namely SynthRef-YouTube-VIS. 
Our experiments show that pretraining a model using our synthetic referring expressions increases its capability to generalize on new data, which is very important in scenarios where training data are not available for a target dataset. Our method, that does not involve any human annotation cost, can be applied to other existing datasets and tasks (\eg object detection or text-to-image retrieval). 
We invite the community to explore further possibilities and benefits out of it.
\\
\section*{Acknowledgments}
\small{This work was partially supported by the EU Erasmus+ programme, and the Spanish Ministry of Economy and Competitivity under grants TEC2016-75976-R (UPC) and RTI2018-095232-B-C22 (UOC).
We gratefully acknowledge the support of NVIDIA Corporation with the donation of GPUs used in this work.
The authors would like to thank Yannis Kalantidis, Konstantinos Karantzalos and the anonymous reviewers for their valuable feedback.
}

\bibliography{anthology,custom}
\bibliographystyle{acl_natbib}




\end{document}